\documentclass[sigconf]{acmart}

\copyrightyear{2020}
\acmYear{2020}
\setcopyright{acmlicensed}\acmConference[MM '20]{Proceedings of the 28th ACM International Conference on Multimedia}{October 12--16, 2020}{Seattle, WA, USA}
\acmBooktitle{Proceedings of the 28th ACM International Conference on Multimedia (MM '20), October 12--16, 2020, Seattle, WA, USA}
\acmPrice{15.00}
\acmDOI{10.1145/3394171.3413532}
\acmISBN{978-1-4503-7988-5/20/10}
\usepackage{multirow}
\begin{document}
\fancyhead{}

\title{A Lip Sync Expert Is All You Need for Speech to Lip Generation In The Wild}

\author{K R Prajwal}
\authornote{Both of the authors have contributed equally to this research.}
\email{prajwal.k@research.iiit.ac.in}
\affiliation{%
  \institution{IIIT, Hyderabad, India}
  %\streetaddress{P.O. Box 1212}
  %\city{Hyderabad}
  %\country{India}
}

\author{Rudrabha Mukhopadhyay}
\authornotemark[1]
\email{radrabha.m@research.iiit.ac.in}
\affiliation{%
  \institution{IIIT, Hyderabad, India}
  %\streetaddress{P.O. Box 1212}
  %\city{Hyderabad}
  %\country{India}
}

\author{Vinay P. Namboodiri}
\email{vpn22@bath.ac.uk}
\affiliation{%
  \institution{University of Bath, England}
  %\streetaddress{P.O. Box 1212}
  %\city{Bath}
  %\country{England}
}

\author{C V Jawahar}
\email{jawahar@iiit.ac.in}
\affiliation{%
  \institution{IIIT, Hyderabad, India}
  %\streetaddress{P.O. Box 1212}
  %\city{Hyderabad}
  %\country{India}
}

\renewcommand{\shortauthors}{Prajwal and Rudrabha, et al.}

\begin{abstract}
  In this work, we investigate the problem of lip-syncing a talking face video of an arbitrary identity to match a target speech segment. Current works excel at producing accurate lip movements on a static image or videos of specific people seen during the training phase. However, they fail to accurately morph the lip movements of arbitrary identities in dynamic, unconstrained talking face videos, resulting in significant parts of the video being out-of-sync with the new audio. We identify key reasons pertaining to this and hence resolve them by learning from a powerful lip-sync discriminator. Next, we propose new, rigorous evaluation benchmarks and metrics to accurately measure lip synchronization in unconstrained videos. Extensive quantitative evaluations on our challenging benchmarks show that the lip-sync accuracy of the videos generated by our Wav2Lip model is almost as good as real synced videos. We provide a demo video clearly showing the substantial impact of our Wav2Lip model and evaluation benchmarks on our website: \url{cvit.iiit.ac.in/research/projects/cvit-projects/a-lip-sync-expert-is-all-you-need-for-speech-to-lip-generation-in-the-wild}. The code and models are released here: \url{github.com/Rudrabha/Wav2Lip}. You can also try out the interactive demo at this link: \url{bhaasha.iiit.ac.in/lipsync}.
\end{abstract}




\begin{teaserfigure}
 \includegraphics[width=\textwidth]{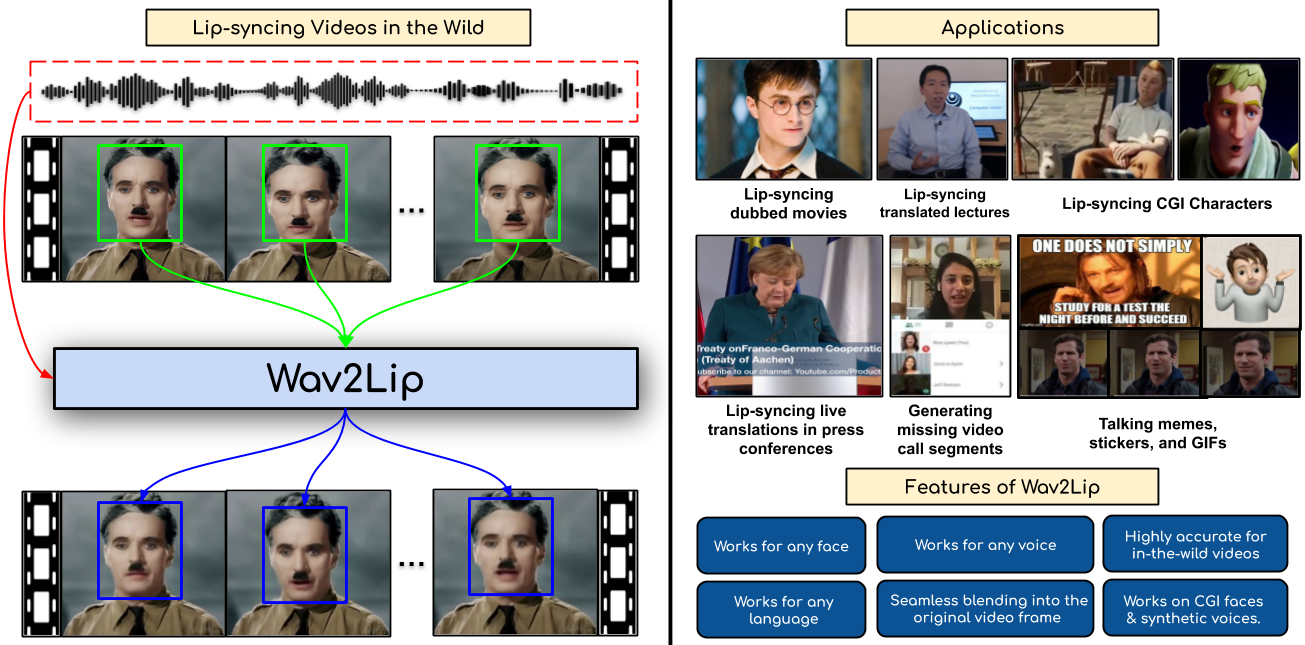}
 \caption{Our novel Wav2Lip model produces significantly more accurate lip-synchronization in dynamic, unconstrained talking face videos. Quantitative metrics indicate that the lip-sync in our generated videos are almost as good as real-synced videos. Thus, we believe that our model can enable a wide range of real-world applications where previous speaker-independent lip-syncing approaches~\cite{jamaludin2019you,kr2019towards} struggle to produce satisfactory results.}
 \label{fig:teaser}
\end{teaserfigure}

\maketitle

\section{Introduction}
With the exponential rise in the consumption of audio-visual content~\cite{videocalling}, rapid video content creation has become a quintessential need. At the same time, making these videos accessible in different languages is also a key challenge. For instance, a deep learning lecture series, a famous movie, or a public address to the nation, if translated to desired target languages, can become accessible to millions of new viewers. A crucial aspect of translating such talking face videos or creating new ones is correcting the lip sync to match the desired target speech. Consequently, lip-syncing talking face videos to match a given input audio stream has received considerable attention~\cite{jamaludin2019you,fried2019text,chen2019sound,thies2019neural,kr2019towards} in the research community. 

Initial works~\cite{kumar2017obamanet,suwajanakorn2017synthesizing} using deep learning in this space learned a mapping from speech representations to lip landmarks using several hours of a single speaker. More recent works~\cite{fried2019text,thies2019neural} in this line directly generate images from speech representations and show exceptional generation quality for specific speakers which they have been trained upon. Numerous practical applications, however, require models that can readily work for generic identities and speech inputs. This has led to the creation of speaker-independent speech to lip generation models~\cite{jamaludin2019you,kr2019towards} that are trained on thousands of identities and voices. They can generate accurate lip motion on a single, static image of any identity in any voice, including that of a synthetic speech generated by a text-to-speech system~\cite{kr2019towards}. However, to be used for applications like translating a lecture/TV series, for example, these models need to be able to morph the broad diversity of lip shapes present in these dynamic, unconstrained videos as well, and not just on static images. 

Our work builds upon this latter class of speaker-independent works that aspire to lip-sync talking face videos of any identity and voice. We find that these models that work well for static images are unable to accurately morph the large variety of lip shapes in unconstrained video content, leading to significant portions of the generated video being out-of-sync with the new target audio. A viewer can recognize an out-of-sync video segment as small as just $\approx 0.05 - 0.1$ seconds~\cite{Chung16a} in duration. Thus, convincingly lip-syncing a real-world video to an entirely new speech is quite challenging, given the tiny degree of allowed error. Further, the fact that we are aiming for a speaker-independent approach without any additional speaker-specific data overhead makes our task even more difficult. Real-world videos contain rapid pose, scale, and illumination changes and the generated face result must also seamlessly blend into the original target video. 

We start by inspecting the existing speaker-independent approaches for speech to lip generation. We find that these models do not adequately penalize wrong lip shapes, either as a result of using only reconstruction losses or weak lip-sync discriminators. We adapt a powerful lip-sync discriminator that can enforce the generator to consistently produce accurate, realistic lip motion. Next, we re-examine the current evaluation protocols and devise new, rigorous evaluation benchmarks derived from three standard test sets. We also propose reliable evaluation metrics using SyncNet~\cite{Chung16a} to precisely evaluate lip sync in unconstrained videos. We also collect and release ReSyncED, a challenging set of real-world videos that can benchmark how the models will perform in practice. We conduct extensive quantitative and subjective human evaluations and outperform previous methods by a large margin across all benchmarks. Our key contributions/claims are as follows:

\begin{itemize}
    \item We propose a novel lip-synchronization network, Wav2Lip, that is significantly more accurate than previous works for lip-syncing arbitrary talking face videos in the wild with arbitrary speech.
    \item We propose a new evaluation framework, consisting of new benchmarks and metrics, to enable a fair judgment of lip synchronization in unconstrained videos.
    \item We collect and release ReSyncED, a Real-world lip-Sync Evaluation Dataset to benchmark the performance of the lip-sync models on completely unseen videos in the wild. 
    \item Wav2Lip is the first speaker-independent model to generate videos with lip-sync accuracy that matches the real synced videos. Human evaluations indicate that the generated videos of Wav2Lip are preferred over existing methods and unsynced versions more than 90\% of the time. 
\end{itemize}

A demo video can be found on our website\footnote{\url{cvit.iiit.ac.in/research/projects/cvit-projects/a-lip-sync-expert-is-all-you-need-for-speech-to-lip-generation-in-the-wild}} with several qualitative examples that clearly illustrate the impact of our model. We will also release an interactive demo on the website allowing users to try out the model using audio and video samples of their choice. The rest of the paper is organized as follows: Section~\ref{section:related} surveys the recent developments in the area of speech to lip generation, Section~\ref{section:arch} discusses the issues with the existing works and describes our proposed approach to mitigate them, Section~\ref{section:quanteval} proposes a new, reliable evaluation framework. We describe the various potential applications and address some of the ethical concerns in Section~\ref{sec:apps} and conclude in Section~\ref{section:conclusion}.

\section{Related Work}
\label{section:related}
\subsection{Constrained Talking Face Generation from Speech}
We first review works on talking face generation that are either constrained by the range of identities they can generate or the range of vocabulary they are limited to. Realistic generation of talking face videos was achieved by a few recent works~\cite{suwajanakorn2017synthesizing,kumar2017obamanet} on videos of Barack Obama. They learn a mapping between the input audio and the corresponding lip landmarks. As they are trained on only a specific speaker, they cannot synthesize for new identities or voices. They also require a large amount of data of a particular speaker, typically a few hours. A recent work along this line~\cite{fried2019text} proposes to seamlessly edit videos of individual speakers by adding or removing phrases from the speech. They still require an hour of data per speaker to achieve this task. Very recently, another work~\cite{thies2019neural} tries to minimize this data overhead by using a two-stage approach, where they first learn speaker-independent features and then learn a rendering mapping with $\approx5$ minutes of data of the desired speaker. However, they train their speaker-independent network on a significantly smaller corpus and also have an additional overhead of requiring clean training data of each target speaker to generate for that speaker. Another limitation of existing works is in terms of the vocabulary. Several works~\cite{zhou2018talking,vougioukas2019realistic,chen2019hierarchical} train on datasets with a limited set of words such as GRID~\cite{cooke2006audio} (56 words), TIMIT~\cite{harte2015tcd} and LRW~\cite{chung2016lip} (1000 words) which significantly hampers a model from learning the vast diversity of phoneme-viseme mappings in real videos~\cite{kr2019towards}. Our work focuses on lip-syncing unconstrained talking face videos to match any target speech, not limited by identities, voices, or vocabulary.

\subsection{Unconstrained Talking Face Generation from Speech}
\label{subsection:unconstrained}
Despite the rise in the number of works on speech-driven face generation, surprisingly, very few works have been designed to lip-sync videos of arbitrary identities, voices, and languages. They are not trained on a small set of identities or a small vocabulary. This allows them to, at test time, lip-sync random identities for any speech. To the best of our knowledge, only two such prominent works~\cite{jamaludin2019you,kr2019towards} exist in the current literature. Note that~\cite{jamaludin2019you} is an extended version of~\cite{chung2017you}. Both these works~\cite{jamaludin2019you,kr2019towards} formulate the task of learning to lip-sync in the wild as follows: \textit{Given a short speech segment $S$ and a random reference face image $R$, the task of the network is to generate a lip-synced version $L_g$ of the input face that matches the audio}. Additionally, the LipGAN model also inputs the target face with bottom-half masked to act as a pose prior. This was crucial as it allowed the generated face crops to be seamlessly pasted back into the original video without further post-processing. It also trains a discriminator in conjunction with the generator to discriminate in-sync or out-of-sync audio-video pairs. Both these works, however, suffer from a significant limitation: they work very well on static images of arbitrary identities but produce inaccurate lip generation when trying to lip-sync unconstrained videos in the wild. In contrast to the GAN setup used in LipGAN~\cite{kr2019towards}, we use a pre-trained, accurate lip-sync discriminator that is not trained further with the generator. We observe that this is an important design choice to achieve much better lip-sync results. 

\section{Accurate Speech-driven Lip-Syncing for Videos in the Wild}
\label{section:arch}
Our core architecture can be summed up as \textit{``Generating accurate lip-sync by learning from a well-trained lip-sync expert"}. To understand this design choice, we first identify two key reasons why existing architectures (section \ref{subsection:unconstrained}) produce inaccurate lip-sync for videos in the wild. We argue that the loss functions, namely the L1 reconstruction loss used in both the existing works~\cite{kr2019towards,jamaludin2019you} and the discriminator loss in LipGAN~\cite{kr2019towards} are inadequate to penalize inaccurate lip-sync generation.

\begin{figure*}
 \includegraphics[width=\textwidth]{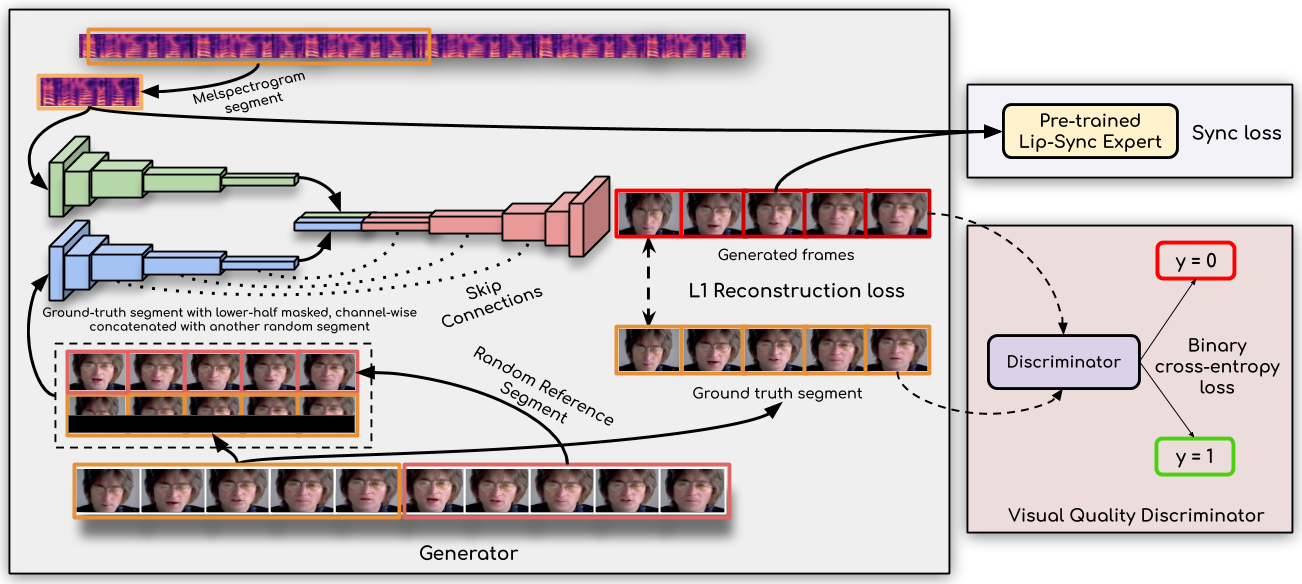}
 
 \caption{Our approach generates accurate lip-sync by learning from an ``already well-trained lip-sync expert". Unlike previous works that employ only a reconstruction loss~\cite{jamaludin2019you} or train a discriminator in a GAN setup~\cite{kr2019towards}, we use a pre-trained discriminator that is already quite accurate at detecting lip-sync errors. We show that fine-tuning it further on the noisy generated faces hampers the discriminator's ability to measure lip-sync, thus also affecting the generated lip shapes. Additionally, we also employ a visual quality discriminator to improve the visual quality along with the sync accuracy.}
 \label{fig:full_arch}
\end{figure*}
\subsection{Pixel-level Reconstruction loss is a Weak Judge of Lip-sync}
The face reconstruction loss is computed for the whole image, to ensure correct pose generation, preservation of identity, and even background around the face. The lip region corresponds to less than 4\% of the total reconstruction loss (based on the spatial extent), so a lot of surrounding image reconstruction is first optimized before the network starts to perform fine-grained lip shape correction. This is further supported by the fact that the network begins morphing lips only at around half-way ($\approx 11^{th}$ epoch) through its training process ($\approx20$ epochs~\cite{kr2019towards}). Thus, it is crucial to have an additional discriminator to judge lip-sync, as also done in LipGAN~\cite{kr2019towards}. But, how powerful is the discriminator employed in LipGAN?

\subsection{A Weak Lip-sync Discriminator}
\label{section:nodisc}
We find that the LipGAN's lip-sync discriminator is only about 56\% accurate while detecting off-sync audio-lip pairs on the LRS2 test set. For comparison, the expert discriminator that we will use in this work is 91\% accurate on the same test set. We hypothesize two major reasons for this difference. Firstly, LipGAN's discriminator uses a single frame to check for lip-sync. In Table~\ref{tab:disc_ablation}, we show that a small temporal context is very helpful while detecting lip-sync. Secondly, the generated images during training contain a lot of artifacts due to the large scale and pose variations. We argue that training the discriminator in a GAN setup on these noisy generated images, as done in LipGAN, results in the discriminator focusing on the visual artifacts instead of the audio-lip correspondence. This leads to a large drop in off-sync detection accuracy (Table~\ref{tab:disc_ablation}). We argue and show that the ``real", accurate concept of lip-sync captured from the actual video frames can be used to accurately discriminate and enforce lip-sync in the generated images.

\subsection{A Lip-sync Expert Is All You Need}
\label{subsection:syncnet}
Based on the above two findings, we propose to use a pre-trained expert lip-sync discriminator that is accurate in detecting sync in real videos. Also, it should not be fine-tuned further on the generated frames like it is done in LipGAN. One such network that has been used to correct lip-sync errors for creating large lip-sync datasets~\cite{Afouras18c,afouras2018lrs3} is the SyncNet~\cite{Chung16a} model. We propose to adapt and train a modified version of SyncNet~\cite{Chung16a} for our task. 

\subsubsection{Overview of SyncNet} SyncNet~\cite{Chung16a} inputs a window $V$ of $T_v$ consecutive face frames (lower half only) and a speech segment $S$ of size $T_a \times D$, where $T_v$ and $T_a$ are the video and audio time-steps respectively. It is trained to discriminate sync between audio and video by randomly sampling an audio window $T_a \times D$ that is either aligned with the video (in-sync) or from a different time-step (out-of-sync). It contains a face encoder and an audio encoder, both comprising of a stack of 2D-convolutions. L2 distance is computed between the embeddings generated from these encoders, and the model is trained with a max-margin loss to minimize (or maximize) the distance between synced (or unsynced) pairs. 

\subsubsection{Our expert lip-sync discriminator} We make the following changes to SyncNet~\cite{Chung16a} to train an expert lip-sync discriminator that suits our lip generation task. Firstly, instead of feeding gray-scale images concatenated channel-wise as in the original model, we feed color images. Secondly, our model is significantly deeper, with residual skip connections~\cite{he2016deep}. Thirdly, inspired by this public implementation\footnote{github.com/joonson/syncnet\_trainer}, we use a different loss function: cosine-similarity with binary cross-entropy loss. That is, we compute a dot product between the ReLU-activated video and speech embeddings $v, s$ to yield a single value between $[0, 1]$ for each sample that indicates the probability that the input audio-video pair is in sync: 

\begin{equation}
    P_{\mathrm{sync}} = \dfrac{v \cdot s}{max(\lVert v \rVert_{2} \cdot \lVert s \rVert_{2}, \epsilon)}
\label{eq:cosine}
\end{equation}

We train our expert lip-sync discriminator on the LRS2 train split ($\approx29$ hours) with a batch size of 64, with $T_v = 5$ frames using the Adam optimizer~\cite{duchi2011adaptive} with an initial learning rate of $1e^{-3}$. Our expert lip-sync discriminator is about 91\% accurate on the LRS2 test set, while the discriminator used in LipGAN is only 56\% accurate on the same test set. 

\subsection{Generating Accurate Lip-sync by learning from a Lip-sync Expert}
Now that we have an accurate lip-sync discriminator, we can now use it to penalize the generator (Figure \ref{fig:full_arch}) for inaccurate generation during the training time. We start by describing the generator architecture. 

\subsubsection{Generator Architecture Details} We use a similar generator architecture as used by LipGAN~\cite{kr2019towards}. Our key contribution lies in training this with the expert discriminator. The generator $G$ contains three blocks: (i) Identity Encoder, (ii) Speech Encoder, and a (iii) Face Decoder. The Identity Encoder is a stack of residual convolutional layers that encode a random reference frame $R$, concatenated with a pose-prior $P$ (target-face with lower-half masked) along the channel axis. The Speech Encoder is also a stack of 2D-convolutions to encode the input speech segment $S$ which is then concatenated with the face representation. The decoder is also a stack of convolutional layers, along with transpose convolutions for upsampling. The generator is trained to minimize L1 reconstruction loss between the generated frames $L_g$ and ground-truth frames $L_G$:

\begin{equation}
L_{\mathrm{recon}} = \frac{1}{N} \sum^{N}_{i=1} ||L_g - L_G||_1
\label{eqn:recon}
\end{equation}

Thus, the generator is similar to the previous works, a 2D-CNN encoder-decoder network that generates each frame independently. How do we then employ our pre-trained expert lip-sync discriminator that needs a temporal window of $T_v=5$ frames as input?

\subsubsection{Penalizing Inaccurate Lip Generation}
During training, as the expert discriminator trained in section \ref{subsection:syncnet} processes $T_v=5$ contiguous frames at a time, we would also need the generator $G$ to generate all the $T_v=5$ frames. We sample a random contiguous window for the reference frames, to ensure as much temporal consistency of pose, etc. across the $T_v$ window. As our generator processes each frame independently, we stack the time-steps along the batch dimension while feeding the reference frames to get an input shape of $(N \cdot T_v, H, W, 3)$, where N, H, W are the batch-size, height, and width respectively. While feeding the generated frames to the expert discriminator, the time-steps are concatenated along the channel-dimension as was also done during the training of the discriminator. The resulting input shape to the expert discriminator is $(N, H/2, W, 3 \cdot T_v)$, where only the lower half of the generated face is used for discrimination. The generator is also trained to minimize the ``expert sync-loss" $E_\mathrm{sync}$ from the expert discriminator:

\begin{equation}
    E_\mathrm{sync} = \frac{1}{N} \sum^{N}_{i=1} - \log(P^{i}_{\mathrm{sync}})
    \label{eqn:syncloss}
\end{equation}

where $P^{i}_{\mathrm{sync}}$ is calculated according to Equation \ref{eq:cosine}. Note that the expert discriminator's weights remain frozen during the training of the generator. This strong discrimination based purely on the lip-sync concept learned from real videos forces the generator to also achieve realistic lip-sync to minimize the lip-sync loss $E_\mathrm{sync}$. 

\subsection{Generating Photo-realistic Faces}
In our experiments, we observed that using a strong lip-sync discriminator forces the generator to produce accurate lip shapes. However, it sometimes results in the morphed regions to be slightly blurry or contain slight artifacts. To mitigate this minor loss in quality, we train a simple visual quality discriminator in a GAN setup along with the generator. Thus, we have two discriminators, one for sync accuracy and another for better visual quality. The lip-sync discriminator is~\textit{not} trained in a GAN setup for reasons explained in~\ref{section:nodisc}. On the other hand, since the visual quality discriminator does not perform any checks on lip-sync and only penalizes unrealistic face generations, it is trained on the generated faces. 

The discriminator $D$ consists of a stack of convolutional blocks. Each block consists of a convolutional layer followed by a Leaky ReLU activation~\cite{maas2013rectifier}. The discriminator is trained to maximize the objective function $L_{disc}$ (Equation~\ref{eqn:disc}):

\begin{align}
    L_{gen} = \mathbb{E}_{x\sim L_{g}}[log(1 - D(x)]\label{eqn:lgen}\\
    L_{disc} = \mathbb{E}_{x\sim L_{G}}[log(D(x))] + L_{gen}
    \label{eqn:disc}
\end{align}

where $L_{g}$ corresponds to the images from the generator $G$, and $L_{G}$ corresponds to the real images.

The generator minimizes Equation~\ref{eqn:gen_loss}, which is the weighted sum of the reconstruction loss (Equation \ref{eqn:recon}), the synchronization loss (Equation~\ref{eqn:syncloss}) and the adversarial loss $L_{gen}$ (Equation~\ref{eqn:lgen}):

\begin{equation}
    L_{\mathrm{total}} = (1 - s_w - s_g) \cdot L_{\mathrm{recon}} + s_w \cdot E_\mathrm{sync} + s_g \cdot L_{gen}
    \label{eqn:gen_loss}
\end{equation}

where $s_w$ is the synchronization penalty weight, $s_g$ is the adversarial loss which are empirically set to $0.03$ and $0.07$ in all our experiments. Thus, our complete network is optimized for both superior sync-accuracy and quality using two disjoint discriminators.  

We train our model only on the LRS2 train set~\cite{Afouras18c}, with a batch size of $80$. We use the Adam optimizer~\cite{duchi2011adaptive} with an initial learning rate of $1e^{-4}$ and betas $\beta_1 = 0.5, \beta_2 = 0.999$ for both the generator and visual quality discriminator $D$. Note that the lip-sync discriminator is not fine-tuned further, so its weights are frozen. We conclude the description of our proposed architecture by explaining how it works during the inference on real videos. Similar to LipGAN~\cite{kr2019towards}, the model generates a talking face video frame-by-frame. The visual input at each time-step is the current face crop (from the source frame), concatenated with the same current face crop with lower-half masked to be used as a pose prior. Thus, during inference, the model does not need to change the pose, significantly reducing artifacts. The corresponding audio segment is also given as input to the speech sub-network, and the network generates the input face crop, but with the mouth region morphed. 

All our code and models will be released publicly. We will now quantitatively evaluate our novel approach against previous models.

\section{Quantitative Evaluation}
\label{section:quanteval}
Despite training only on the LRS2 train set, we evaluate our model across $3$ different datasets. But before doing so, we re-investigate the current evaluation framework followed in prior works and why it is far from being an ideal way to evaluate works in this space.

\begin{table*}[ht]
  \setlength{\tabcolsep}{4pt}
    \centering
    
    \begin{tabular}{|l||c|c|c||c|c|c||c|c|c|}
    \hline
    & \multicolumn{3}{c||}{LRW~\cite{chung2016lip}} & \multicolumn{3}{c||}{LRS2~\cite{Afouras18c}} & \multicolumn{3}{c|}{LRS3~\cite{afouras2018lrs3}}
     \\
    \hline
    \textbf{Method} & \textbf{LSE-D	$\downarrow$} & \textbf{LSE-C $\uparrow$} & \textbf{FID $\downarrow$} & \textbf{LSE-D $\downarrow$} & \textbf{LSE-C $\uparrow$} & \textbf{FID $\downarrow$} & \textbf{LSE-D $\downarrow$} & \textbf{LSE-C $\uparrow$} & \textbf{FID	$\downarrow$} \\
    \hline
    Speech2Vid~\cite{jamaludin2019you} & 13.14 & 1.762 & 11.15 & 14.23 & 1.587 & 12.32 & 13.97 & 1.681 & 11.91\\
    LipGAN~\cite{kr2019towards} & 10.05 & 3.350 & 2.833 & 10.33 & 3.199 & 4.861 & 10.65 & 3.193 & 4.732 \\
    \textbf{Wav2Lip (ours)} & \textbf{6.512} & \textbf{7.490} & 3.189 & \textbf{6.386} & \textbf{7.789} & 4.887 & \textbf{6.652} & \textbf{7.887} & 4.844 \\
    \textbf{Wav2Lip + GAN (ours)} & 6.774 & 7.263 & \textbf{2.475} & 6.469 & 7.781 & \textbf{4.446} & 6.986 & 7.574 & \textbf{4.350} \\
    \hline
    Real Videos & 7.012  & 6.931  & --- & 6.736 & 7.838 & --- & 6.956 & 7.592 & --- \\
    \hline
    
    \end{tabular}
    \caption{We propose two new metrics ``Lip-Sync Error-Distance" (lower is better) and ``Lip-Sync Error-Confidence" (higher is better), that can reliably measure the lip-sync accuracy in unconstrained videos. We see that the lip-sync accuracy of the videos generated using Wav2Lip is almost as good as real synced videos. Note that we only train on the train set on LRS2~\cite{Afouras18c}, but we comfortably generalize across all datasets without any further fine-tuning. We also report the FID score (lower is better), which clearly shows that using a visual quality discriminator improves the quality by a significant margin.}
    \vspace{-0.7cm}
    \label{tab:lse_scores}
\end{table*}

\subsection{Re-thinking the Evaluation Framework for Speech-driven Lip-Syncing in the Wild}
The current evaluation framework for speaker-independent lip-syncing judges the models differently from how it is used while lip-syncing a real video. Specifically, instead of feeding the current frame as a reference (as described in the previous section), a random frame in the video is chosen as the reference to not leak the correct lip information during evaluation. We strongly argue that the evaluation framework in the previous paragraph is not ideal for evaluating the lip-sync quality and accuracy. Upon a closer examination of the above-mentioned evaluation system, we observed a few key limitations, which we discuss below. 

\subsubsection{Does not reflect the real-world usage} As discussed before, during generation at test time, the model must not change the pose, as the generated face needs to be seamlessly pasted into the frame. However, the current evaluation framework feeds random reference frames in the input, thus demanding the network to change the pose. Thus, the above system does not evaluate how the model would be used in the real world.

\subsubsection{Inconsistent evaluation} As the reference frames are chosen at random, this means the test data is not consistent across different works. This would lead to an unfair comparison and hinder the reproducibility of results.

\subsubsection{Does not support checking for temporal consistency}
As the reference frames are randomly chosen at each time-step, temporal consistency is already lost as the frames are generated at random poses and scales. The current framework cannot support a new metric or a future method that aims to study the temporal consistency aspect of this problem. 

\subsubsection{Current metrics are not specific to lip-sync}
The existing metrics, such as SSIM~\cite{wang2004image} and PSNR, were developed to evaluate overall image quality and not fine-grained lip-sync errors. Although LMD~\cite{chen2018lip} focuses on the lip region, we found that lip landmarks can be quite inaccurate on generated faces. Thus, there is a need for a metric that is designed specifically for measuring lip-sync errors. 

\subsection{A Novel Benchmark and Metric for Evaluating Lip-Sync in the Wild}

The reason for sampling random frames for evaluation is because, the current frame is already in sync with the speech, leading to leakage of lip-shape in the input itself. And previous works have not tried sampling different speech segments instead of sampling a different frame, as the ground-truth lip shape for the sampled speech is unavailable. 

\subsubsection{A Metric to Measure the Lip-Sync Error}
We propose to use the pre-trained SyncNet~\cite{Chung16a} available publicly\footnote{github.com/joonson/syncnet\_python} to measure the lip-sync error between the generated frames and the randomly chosen speech segment. The accuracy of SyncNet averaged over a video clip is over 99\%~\cite{Chung16a}. Thus, we believe this can be a good automatic evaluation method that explicitly tests for accurate lip-sync in unconstrained videos in the wild. Note that this is not the expert lip-sync discriminator that we have trained above, but the one released by~\citet{Chung16a}, which was trained on a different, non-public dataset. Using a SyncNet resolves major issues of the existing evaluation framework. We no longer need to sample random, temporally incoherent frames and SyncNet also takes into account short-range temporal consistency while evaluating lip-sync. Thus, we propose two new metrics automatically determined using the SyncNet model. The first is the average error measure calculated in terms of the distance between the lip and audio representations, which we code-name as ``LSE-D" (``Lip Sync Error - Distance"). A lower LSE-D denotes a higher audio-visual match, i.e., the speech and lip movements are in sync. The second metric is the average confidence score, which we code-name as ``LSE-C" (Lip Sync Error - Confidence). Higher the confidence, the better the audio-video correlation. A lower confidence score denotes that there are several portions of the video with completely out-of-sync lip movements. Further details can be found in the SyncNet paper~\cite{Chung16a}.

\begin{table*}[ht]
  \setlength{\tabcolsep}{4pt}
    \centering
    
    \begin{tabular}{|l|c|c|c|c||c|c|c|c|}
    \hline
    
    \textbf{Method} & \textbf{Video Type} & \textbf{LSE-D $\downarrow$} & \textbf{LSE-C $\uparrow$} & \textbf{FID $\downarrow$} & \textbf{Sync Acc.} & \textbf{Visual Qual.} & \textbf{Overall Experience} & \textbf{Preference}\\
    \hline
    Unsynced Orig. Videos & \multirow{5}{*}{Dubbed} & 12.63 & 0.896 & --- & 0.21 & 4.81 & 3.07 & 3.15\% \\
    Speech2Vid~\cite{jamaludin2019you} & & 14.76 & 1.121 & 19.31 & 1.14 & 0.93 & 0.84 & 0.00\% \\
    LipGAN~\cite{kr2019towards} & & 10.61 & 2.857 & 12.87 & 2.98 & 3.91 & 3.45 & 2.35\%\\
    \textbf{Wav2Lip (ours)} & & \textbf{6.843} & \textbf{7.265} & 15.65 & \textbf{4.13} & 3.87 & 4.04 & 34.3\%\\
    \textbf{Wav2Lip + GAN (ours)} & & 7.318 & 6.851 & \textbf{11.84} & 4.08 & \textbf{4.12} & \textbf{4.13} & \textbf{60.2\%}\\
    \hline
    Without Lip-syncing & \multirow{5}{*}{Random} & 17.12 & 2.014 & --- & 0.15 & 4.56 & 2.98 & 3.24\%\\
    Speech2Vid~\cite{jamaludin2019you} & & 15.22 & 1.086 & 19.98 & 0.87 & 0.79 & 0.73 & 0.00\% \\
    LipGAN~\cite{kr2019towards} & & 11.01 & 3.341 & 14.60 & 3.42 & 3.77 & 3.57 & 3.16\%\\
    \textbf{Wav2Lip (ours)} & & \textbf{6.691} & \textbf{8.220} & 14.47 & \textbf{4.24} & 3.68 & 4.01 & 29.1\%\\
    \textbf{Wav2Lip + GAN (ours)} & & 7.066 & 8.011 & \textbf{13.12} & 4.18 & \textbf{4.05} & \textbf{4.15} & \textbf{64.5\%}\\
    \hline
    Without Lip-syncing & \multirow{6}{*}{TTS} & 16.89 & 2.557 & --- & 0.11 & 4.67 & 3.32 & 8.32\% \\
    Speech2Vid~\cite{jamaludin2019you} & & 14.39 & 1.471 & 17.96 & 0.76 & 0.71 & 0.69 & 0.00\%\\
    LipGAN~\cite{kr2019towards} & & 10.90 & 3.279 & 11.91 & 2.87 & 3.69 & 3.14 & 1.64\%\\
    \textbf{Wav2Lip (ours)} & & \textbf{6.659} & \textbf{8.126} & 12.77 & \textbf{3.98} & 3.87 & 3.92 & 41.2\%\\
    \textbf{Wav2Lip + GAN (ours)} & & 7.225 & 7.651 & \textbf{11.15} & 3.85 & \textbf{4.13} & \textbf{4.05} & \textbf{51.2\%}\\
    Untranslated Videos & & 7.767 & 7.047 & --- & 4.83 & 4.91 & --- & --- \\
    \hline
    \end{tabular}
    \caption{Real world evaluation using our newly collected ReSyncED benchmark. We evaluate using both quantitative metrics and human evaluation scores across three classes of real videos. We can see that in all cases, the Wav2Lip model produces high-quality, accurate lip-syncing videos. Specifically, the metrics indicate that our lip-synced videos are as good as the real synced videos. We also note that human evaluations indicate that there is a scope for improvement when trying to lip-sync TTS generated speech. Finally, it is worth noting that our lip-synced videos are preferred over existing methods or the actual unsynced videos over 90\% of the time.}
    \vspace{-0.7cm}
    \label{tab:realvids}
\end{table*}

\subsubsection{A Consistent Benchmark to Evaluate Lip-sync in the wild}
\label{subsec:consistent}
Now that we have an automatic, reliable metric that can be computed for any video and audio pairs, we can sample random speech samples instead of a random frame at each time-step. Thus, we can create a list of pairs of video and a pseudo-randomly chosen audio as a consistent test set. We create three consistent benchmarks test sets, one each using the test set videos of LRS2~\cite{Afouras18c}, LRW~\cite{chung2016lip}, and LRS3~\cite{afouras2018lrs3} respectively. For each video $V_s$, we take the audio from another randomly-sampled video $V_t$ with the condition that the length of the speech $V_t$ be less than $V_s$. We create $14K$ audio-video pairs using LRS2. Using the LRW test set, we create $28K$ pairs, and this set measures the performance on frontal/near-frontal videos~\cite{Afouras18}. We also create $14K$ pairs using the LRS3 test set, which will be a benchmark for lip-syncing in profile views as well. The complete evaluation toolkit will be publicly released for consistent and reliable benchmarking of lip-syncing videos in the wild. 

\subsection{Comparing the Models on the New Benchmark}
We compare the previous two approaches~\cite{kr2019towards,jamaludin2019you} on our newly created test set using the LSE-D and LSE-C metrics. During inference, we now feed the same reference and pose-prior at each time-step, similar to how it has been described before in the architecture section. The mean LSE-D and LSE-C scores are shown in Table \ref{tab:lse_scores} for the audio-video pairs in all three test splits. Additionally, to measure the quality of the generated faces, we also report the Fréchet Inception Distance (FID). Our method outperforms previous approaches by a large margin indicating the significant effect of strong lip-sync discrimination. We can also see the significant improvement in quality after using a visual quality discriminator along with a lip-sync expert discriminator. However, we observe a minor drop in sync accuracy after using the visual quality discriminator. Thus, we will release both of these models, as they have a slight trade-off between visual quality and sync accuracy.

\subsection{Real-World Evaluation}
Apart from evaluating on just the standard datasets, our new evaluation framework and metrics allow us to evaluate on real-world videos on which these models are most likely to be used. Further, given the sensitivity of humans to audio-lip synchronization~\cite{Chung16a}, it is necessary to also evaluate our results with the help of human evaluators. Thus, contrary to the previous works on speaker-independent lip-syncing, we conduct both quantitative and human evaluation experiments on unconstrained real videos from the web for the first time. Thus, we collect and publicly release ``ReSyncED" a ``Real-world Evaluation Dataset" to subjectively and objectively benchmark the performance of lip-sync works. 

\subsubsection{Curating ReSyncED} All our videos are downloaded from YouTube. We specifically choose three types of video examples. The first type ``Dubbed", contains videos where the audio is naturally out-of-sync, such as dubbed movie clips or public addresses that are live translated to a different language (so the addresser's lips are out-of-sync with the translated speech). The second type is ``Random", where we have a collection of videos and we create random audio-visual pairs similar to~\ref{subsec:consistent}. The third and final type of videos, ``TTS", has been specifically chosen for benchmarking the lip-syncing performance on synthetic speech obtained from a text-to-speech system. This is essential for future works that aspire to automatically translate videos (Face-to-Face Translation~\cite{kr2019towards}) or rapidly create new video content. We manually transcribe the text, use Google Translate (about $5$ languages totally) and publicly available text-to-speech models to generate synthetic translated speech for the videos in this category. The task is to correct lip movements in the original videos to match this synthetic speech.

\subsubsection{Real-world Evaluation on ReSyncED}
\begin{figure*}
 \includegraphics[width=\textwidth]{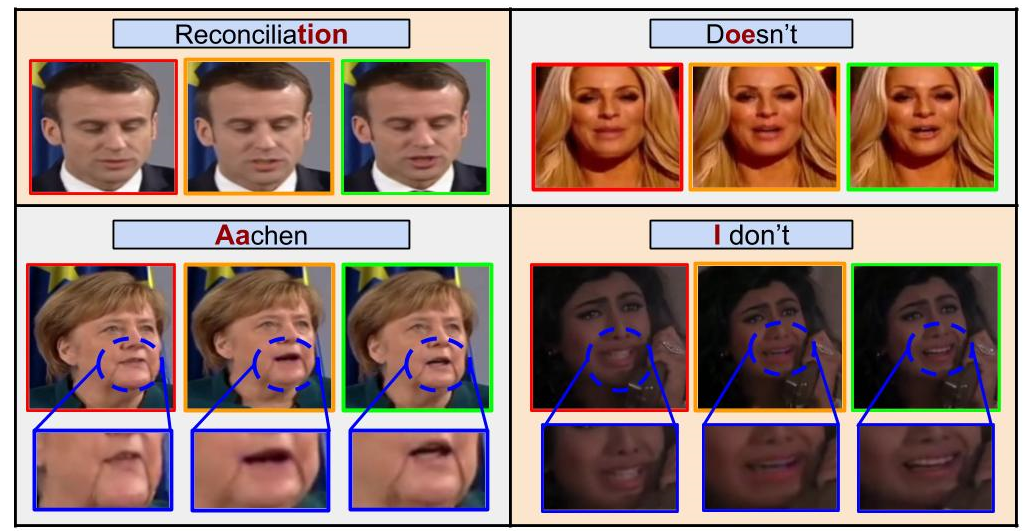}
 \vspace{-0.6cm}
 \caption{Examples of faces generated from our proposed models (green and yellow outlines). We compare with the current best approach~\cite{kr2019towards} (red outline). The text is shown for illustration to denote the utterance being spoken in the frame shown. We can see that our model produces accurate, natural lip shapes. The addition of a visual quality discriminator also significantly improves the visual quality. We strongly encourage the reader to check out the demo video on our website.}
 \label{fig:qual_results}
\end{figure*}

We first evaluate the generated real video results using our new automatic metrics, ``LSE-D" and ``LSE-C" obtained from SyncNet~\cite{Chung16a}. For the human evaluation, we ask $14$ evaluators to judge the different synced versions of the videos based on the following parameters: (a) Sync Accuracy (b) Visual Quality (to evaluate the extent of visual artifacts), (c) Overall Experience (to evaluate the  overall experience of the audio-visual content), and (d) Preference, where the viewer chooses the version of the video that is most appealing to watch. The first three parameters are scored between $1 - 5$, and (d) is a single-choice voting, and we report the percentage of votes obtained by a model. We evaluate each of the three classes of videos separately and report our results in Table~\ref{tab:realvids}. An outcome worth noting is that the previous works~\cite{kr2019towards,jamaludin2019you} which produce several out-of-sync segments are less preferred over the unsynced version as the latter still preserves good Visual quality. Thus, ours is the first work that provides a significant improvement over unsynced talking face videos in-the-wild. We also show some qualitative comparisons in Figure~\ref{fig:qual_results} which contains a few generated samples from the ReSyncED test set.

\subsection{Is our expert discriminator best among the alternatives?}
\begin{table}[ht]
  \begin{tabular}{|l|c|c||c|c|}
    \hline
    Model & Fine-tuned? & Off-sync Acc. & LSE-D & LSE-C \\
    \hline
    $T_v=1$~\cite{kr2019towards} & \checkmark & 55.6\% & 10.33 & 3.19\\
    Ours, $T_v=1$ & $\times$ & 79.3\% & 8.583 & 4.845\\
    \hline
    Ours, $T_v=3$ & \checkmark & 72.3\% & 10.14 & 3.214 \\
    Ours, $T_v=3$ & $\times$ & 87.4\% & 7.230 & 6.533\\
    \hline
    Ours, $T_v=5$ & \checkmark & 73.6\% & 9.953 & 3.508\\
    \textbf{Ours, $T_v=5$} & \textbf{$\times$} & \textbf{91.6\%} & \textbf{6.386} & \textbf{7.789}\\
  \hline
 \end{tabular}
  \caption{A larger temporal window allows for better lip-sync discrimination. On the other hand, training the lip-sync discriminator on the generated faces deteriorates its ability to detect off-sync audio-lip pairs. Consequently, training a lip-sync generator using such a discriminator leads to poorly lip-synced videos.}
  \label{tab:disc_ablation}
  \vspace{-0.7cm}
\end{table}
Our expert discriminator uses $T_v=5$ video frames to measure the lip-sync error. It is also not fine-tuned on the generated faces in a GAN setup. We justify these two design choices in this ablation study. We can test the discriminator's performance by randomly sampling in-sync and off-sync pairs from the LRS2 test set. We vary the size of $T_v=1, 3, 5$ to understand its effect on detecting sync. We also fine-tune/freeze each of the three variants of $T_v$ while training the Wav2Lip model. Thus, we get a total of $6$ variations in Table~\ref{tab:disc_ablation} from which we can clearly make two observations. Increasing the temporal window size $T_v$ consistently provides a better lip-sync discrimination performance. More importantly, we see that if we fine-tune the discriminator on the generated faces that contain artifacts, then the discriminator loses its ability to detect out-of-sync audio-visual pairs. We argue that this happens because the fine-tuned discriminator focuses on the visual artifacts in the generated faces for discrimination, rather than the fine-grained audio-lip correspondence. Thus, it classifies the \textit{real} unsynced pairs as ``in-sync", since these real face images do not contain any artifacts. Down the line, using such a weak discriminator leads to poor lip-sync penalization for our generator, resulting in poorly lip-synced talking face videos.

\section{Applications \& Fair Use}
\label{sec:apps}
At a time when our content consumption and social communication is becoming increasingly audio-visual, there is a dire need for large-scale video translation and creation. Wav2Lip can play a vital role in fulfilling these needs, as it is accurate for videos in the wild. For instance, online lecture videos that are typically in English can now be lip-synced to (automatically) dubbed speech in other local languages (Table~\ref{tab:realvids}, last block). We can also lip-sync dubbed movies making them pleasant to watch (Table~\ref{tab:realvids}, first block). Every day throughout the globe, press conferences and public addresses are live translated but the addresser's lips are out of sync with the translated speech. Our model can seamlessly correct this. Automatically animating the lips of CGI characters to the voice actors' speech can save several hours of manual effort while creating animated movies and rich, conversational game content. We demonstrate our model on all these applications and more in the demo video on our website. 

We believe that it is also essential to discuss and promote fair use of the increasingly capable lip-sync works. The vast applicability of our models with near-realistic lip-syncing capabilities for any identity and voice raises concerns about the potential for misuse. Thus, we strongly suggest that any result created using our code and models must unambiguously present itself as synthetic. In addition to the strong positive impact mentioned above, our intention to completely open-source our work is that it can simultaneously also encourage efforts~\cite{Tursman_2020_CVPR_Workshops,Hsu2020DeepFI,tolosana2020deepfakes,dolhansky2020deepfake} in detecting manipulated video content and their misuse. We believe that Wav2Lip can enable several positive applications and also encourage productive discussions and research efforts regarding fair use of synthetic content.

\section{Conclusion}
\label{section:conclusion}
In this work, we proposed a novel approach to generate accurate lip-synced videos in the wild. We have highlighted two major reasons why current approaches are inaccurate while lip-syncing unconstrained talking face videos. Based on this, we argued that a pre-trained, accurate lip-sync ``expert" can enforce accurate, natural lip motion generation. Before evaluating our model, we re-examined the current quantitative evaluation framework and highlight several major issues. To resolve them, we proposed several new evaluation benchmarks and metrics, and also a real-world evaluation set. We believe future works can be reliably judged in this new framework. Our Wav2Lip model outperforms the current approaches by a large margin in both quantitative metrics and human evaluations. We also investigated the reasons behind our design choices in the discriminator in an ablation study. We encourage the readers to view the demo video on our website. We believe our efforts and ideas in this problem can lead to new directions such as synthesizing expressions and head-poses along with the accurate lip movements.

\newpage
\bibliographystyle{ACM-Reference-Format}
\balance
\bibliography{references}


\begin{thebibliography}{28}

%%% the Web address.

\ifx \showCODEN    \undefined \def \showCODEN     #1{\unskip}     \fi
\ifx \showDOI      \undefined \def \showDOI       #1{#1}\fi
\ifx \showISBNx    \undefined \def \showISBNx     #1{\unskip}     \fi
\ifx \showISBNxiii \undefined \def \showISBNxiii  #1{\unskip}     \fi
\ifx \showISSN     \undefined \def \showISSN      #1{\unskip}     \fi
\ifx \showLCCN     \undefined \def \showLCCN      #1{\unskip}     \fi
\ifx \shownote     \undefined \def \shownote      #1{#1}          \fi
\ifx \showarticletitle \undefined \def \showarticletitle #1{#1}   \fi
\ifx \showURL      \undefined \def \showURL       {\relax}        \fi
% invisible to TeX
\providecommand\bibfield[2]{#2}
\providecommand\bibinfo[2]{#2}
\providecommand\natexlab[1]{#1}
\providecommand\showeprint[2][]{arXiv:#2}

\bibitem[\protect\citeauthoryear{Afouras, Chung, Senior, Vinyals, and
  Zisserman}{Afouras et~al\mbox{.}}{2018c}]%
        {Afouras18c}
\bibfield{author}{\bibinfo{person}{T. Afouras}, \bibinfo{person}{J.~S. Chung},
  \bibinfo{person}{A. Senior}, \bibinfo{person}{O. Vinyals}, {and}
  \bibinfo{person}{A. Zisserman}.} \bibinfo{year}{2018}\natexlab{c}.
\newblock \showarticletitle{Deep Audio-Visual Speech Recognition}. In
  \bibinfo{booktitle}{\emph{arXiv:1809.02108}}.
\newblock


\bibitem[\protect\citeauthoryear{Afouras, Chung, and Zisserman}{Afouras
  et~al\mbox{.}}{2018a}]%
        {Afouras18}
\bibfield{author}{\bibinfo{person}{T. Afouras}, \bibinfo{person}{J.~S. Chung},
  {and} \bibinfo{person}{A. Zisserman}.} \bibinfo{year}{2018}\natexlab{a}.
\newblock \showarticletitle{The Conversation: Deep Audio-Visual Speech
  Enhancement}. In \bibinfo{booktitle}{\emph{INTERSPEECH}}.
\newblock


\bibitem[\protect\citeauthoryear{Afouras, Chung, and Zisserman}{Afouras
  et~al\mbox{.}}{2018b}]%
        {afouras2018lrs3}
\bibfield{author}{\bibinfo{person}{Triantafyllos Afouras},
  \bibinfo{person}{Joon~Son Chung}, {and} \bibinfo{person}{Andrew Zisserman}.}
  \bibinfo{year}{2018}\natexlab{b}.
\newblock \showarticletitle{LRS3-TED: a large-scale dataset for visual speech
  recognition}.
\newblock \bibinfo{journal}{\emph{arXiv preprint arXiv:1809.00496}}
  (\bibinfo{year}{2018}).
\newblock


\bibitem[\protect\citeauthoryear{Chen, Li, K~Maddox, Duan, and Xu}{Chen
  et~al\mbox{.}}{2018}]%
        {chen2018lip}
\bibfield{author}{\bibinfo{person}{Lele Chen}, \bibinfo{person}{Zhiheng Li},
  \bibinfo{person}{Ross K~Maddox}, \bibinfo{person}{Zhiyao Duan}, {and}
  \bibinfo{person}{Chenliang Xu}.} \bibinfo{year}{2018}\natexlab{}.
\newblock \showarticletitle{Lip movements generation at a glance}. In
  \bibinfo{booktitle}{\emph{Proceedings of the European Conference on Computer
  Vision (ECCV)}}. \bibinfo{pages}{520--535}.
\newblock


\bibitem[\protect\citeauthoryear{Chen, Maddox, Duan, and Xu}{Chen
  et~al\mbox{.}}{2019a}]%
        {chen2019hierarchical}
\bibfield{author}{\bibinfo{person}{Lele Chen}, \bibinfo{person}{Ross~K Maddox},
  \bibinfo{person}{Zhiyao Duan}, {and} \bibinfo{person}{Chenliang Xu}.}
  \bibinfo{year}{2019}\natexlab{a}.
\newblock \showarticletitle{Hierarchical cross-modal talking face generation
  with dynamic pixel-wise loss}. In \bibinfo{booktitle}{\emph{Proceedings of
  the IEEE Conference on Computer Vision and Pattern Recognition}}.
  \bibinfo{pages}{7832--7841}.
\newblock


\bibitem[\protect\citeauthoryear{Chen, Zheng, Maddox, Duan, and Xu}{Chen
  et~al\mbox{.}}{2019b}]%
        {chen2019sound}
\bibfield{author}{\bibinfo{person}{Lele Chen}, \bibinfo{person}{Haitian Zheng},
  \bibinfo{person}{Ross~K Maddox}, \bibinfo{person}{Zhiyao Duan}, {and}
  \bibinfo{person}{Chenliang Xu}.} \bibinfo{year}{2019}\natexlab{b}.
\newblock \showarticletitle{Sound to Visual: Hierarchical Cross-Modal Talking
  Face Video Generation}. In \bibinfo{booktitle}{\emph{IEEE Computer Society
  Conference on Computer Vision and Pattern Recognition workshops}}.
\newblock


\bibitem[\protect\citeauthoryear{Chung, Jamaludin, and Zisserman}{Chung
  et~al\mbox{.}}{2017}]%
        {chung2017you}
\bibfield{author}{\bibinfo{person}{Joon~Son Chung}, \bibinfo{person}{Amir
  Jamaludin}, {and} \bibinfo{person}{Andrew Zisserman}.}
  \bibinfo{year}{2017}\natexlab{}.
\newblock \showarticletitle{You said that?}
\newblock \bibinfo{journal}{\emph{arXiv preprint arXiv:1705.02966}}
  (\bibinfo{year}{2017}).
\newblock


\bibitem[\protect\citeauthoryear{Chung and Zisserman}{Chung and
  Zisserman}{2016a}]%
        {chung2016lip}
\bibfield{author}{\bibinfo{person}{Joon~Son Chung} {and}
  \bibinfo{person}{Andrew Zisserman}.} \bibinfo{year}{2016}\natexlab{a}.
\newblock \showarticletitle{Lip reading in the wild}. In
  \bibinfo{booktitle}{\emph{Asian Conference on Computer Vision}}. Springer,
  \bibinfo{pages}{87--103}.
\newblock


\bibitem[\protect\citeauthoryear{Chung and Zisserman}{Chung and
  Zisserman}{2016b}]%
        {Chung16a}
\bibfield{author}{\bibinfo{person}{Joon~Son Chung} {and}
  \bibinfo{person}{Andrew Zisserman}.} \bibinfo{year}{2016}\natexlab{b}.
\newblock \showarticletitle{Out of time: automated lip sync in the wild}. In
  \bibinfo{booktitle}{\emph{Workshop on Multi-view Lip-reading, ACCV}}.
\newblock


\bibitem[\protect\citeauthoryear{Cooke, Barker, Cunningham, and Shao}{Cooke
  et~al\mbox{.}}{2006}]%
        {cooke2006audio}
\bibfield{author}{\bibinfo{person}{Martin Cooke}, \bibinfo{person}{Jon Barker},
  \bibinfo{person}{Stuart Cunningham}, {and} \bibinfo{person}{Xu Shao}.}
  \bibinfo{year}{2006}\natexlab{}.
\newblock \showarticletitle{An audio-visual corpus for speech perception and
  automatic speech recognition}.
\newblock \bibinfo{journal}{\emph{The Journal of the Acoustical Society of
  America}} \bibinfo{volume}{120}, \bibinfo{number}{5} (\bibinfo{year}{2006}),
  \bibinfo{pages}{2421--2424}.
\newblock


\bibitem[\protect\citeauthoryear{Dolhansky, Bitton, Pflaum, Lu, Howes, Wang,
  and Ferrer}{Dolhansky et~al\mbox{.}}{2020}]%
        {dolhansky2020deepfake}
\bibfield{author}{\bibinfo{person}{Brian Dolhansky}, \bibinfo{person}{Joanna
  Bitton}, \bibinfo{person}{Ben Pflaum}, \bibinfo{person}{Jikuo Lu},
  \bibinfo{person}{Russ Howes}, \bibinfo{person}{Menglin Wang}, {and}
  \bibinfo{person}{Cristian~Canton Ferrer}.} \bibinfo{year}{2020}\natexlab{}.
\newblock \bibinfo{title}{The DeepFake Detection Challenge Dataset}.
\newblock
\newblock
\showeprint[arxiv]{2006.07397}~[cs.CV]


\bibitem[\protect\citeauthoryear{Duchi, Hazan, and Singer}{Duchi
  et~al\mbox{.}}{2011}]%
        {duchi2011adaptive}
\bibfield{author}{\bibinfo{person}{John Duchi}, \bibinfo{person}{Elad Hazan},
  {and} \bibinfo{person}{Yoram Singer}.} \bibinfo{year}{2011}\natexlab{}.
\newblock \showarticletitle{Adaptive subgradient methods for online learning
  and stochastic optimization.}
\newblock \bibinfo{journal}{\emph{Journal of machine learning research}}
  \bibinfo{volume}{12}, \bibinfo{number}{7} (\bibinfo{year}{2011}).
\newblock


\bibitem[\protect\citeauthoryear{Fried, Tewari, Zollh{\"o}fer, Finkelstein,
  Shechtman, Goldman, Genova, Jin, Theobalt, and Agrawala}{Fried
  et~al\mbox{.}}{2019}]%
        {fried2019text}
\bibfield{author}{\bibinfo{person}{Ohad Fried}, \bibinfo{person}{Ayush Tewari},
  \bibinfo{person}{Michael Zollh{\"o}fer}, \bibinfo{person}{Adam Finkelstein},
  \bibinfo{person}{Eli Shechtman}, \bibinfo{person}{Dan~B Goldman},
  \bibinfo{person}{Kyle Genova}, \bibinfo{person}{Zeyu Jin},
  \bibinfo{person}{Christian Theobalt}, {and} \bibinfo{person}{Maneesh
  Agrawala}.} \bibinfo{year}{2019}\natexlab{}.
\newblock \showarticletitle{Text-based editing of talking-head video}.
\newblock \bibinfo{journal}{\emph{ACM Transactions on Graphics (TOG)}}
  \bibinfo{volume}{38}, \bibinfo{number}{4} (\bibinfo{year}{2019}),
  \bibinfo{pages}{1--14}.
\newblock


\bibitem[\protect\citeauthoryear{Harte and Gillen}{Harte and Gillen}{2015}]%
        {harte2015tcd}
\bibfield{author}{\bibinfo{person}{Naomi Harte} {and} \bibinfo{person}{Eoin
  Gillen}.} \bibinfo{year}{2015}\natexlab{}.
\newblock \showarticletitle{TCD-TIMIT: An audio-visual corpus of continuous
  speech}.
\newblock \bibinfo{journal}{\emph{IEEE Transactions on Multimedia}}
  \bibinfo{volume}{17}, \bibinfo{number}{5} (\bibinfo{year}{2015}),
  \bibinfo{pages}{603--615}.
\newblock


\bibitem[\protect\citeauthoryear{He, Zhang, Ren, and Sun}{He
  et~al\mbox{.}}{2016}]%
        {he2016deep}
\bibfield{author}{\bibinfo{person}{Kaiming He}, \bibinfo{person}{Xiangyu
  Zhang}, \bibinfo{person}{Shaoqing Ren}, {and} \bibinfo{person}{Jian Sun}.}
  \bibinfo{year}{2016}\natexlab{}.
\newblock \showarticletitle{Deep residual learning for image recognition}. In
  \bibinfo{booktitle}{\emph{Proceedings of the IEEE conference on computer
  vision and pattern recognition}}. \bibinfo{pages}{770--778}.
\newblock


\bibitem[\protect\citeauthoryear{Hsu, Zhuang, and Lee}{Hsu
  et~al\mbox{.}}{2020}]%
        {Hsu2020DeepFI}
\bibfield{author}{\bibinfo{person}{Chih-Chung Hsu}, \bibinfo{person}{Yi-Xiu
  Zhuang}, {and} \bibinfo{person}{Chia-Yen Lee}.}
  \bibinfo{year}{2020}\natexlab{}.
\newblock \showarticletitle{Deep Fake Image Detection based on Pairwise
  Learning}.
\newblock \bibinfo{journal}{\emph{Applied Sciences}}  \bibinfo{volume}{10}
  (\bibinfo{year}{2020}), \bibinfo{pages}{370}.
\newblock


\bibitem[\protect\citeauthoryear{Jamaludin, Chung, and Zisserman}{Jamaludin
  et~al\mbox{.}}{2019}]%
        {jamaludin2019you}
\bibfield{author}{\bibinfo{person}{Amir Jamaludin}, \bibinfo{person}{Joon~Son
  Chung}, {and} \bibinfo{person}{Andrew Zisserman}.}
  \bibinfo{year}{2019}\natexlab{}.
\newblock \showarticletitle{You said that?: Synthesising talking faces from
  audio}.
\newblock \bibinfo{journal}{\emph{International Journal of Computer Vision}}
  \bibinfo{volume}{127}, \bibinfo{number}{11-12} (\bibinfo{year}{2019}),
  \bibinfo{pages}{1767--1779}.
\newblock


\bibitem[\protect\citeauthoryear{KR, Mukhopadhyay, Philip, Jha, Namboodiri, and
  Jawahar}{KR et~al\mbox{.}}{2019}]%
        {kr2019towards}
\bibfield{author}{\bibinfo{person}{Prajwal KR}, \bibinfo{person}{Rudrabha
  Mukhopadhyay}, \bibinfo{person}{Jerin Philip}, \bibinfo{person}{Abhishek
  Jha}, \bibinfo{person}{Vinay Namboodiri}, {and} \bibinfo{person}{CV
  Jawahar}.} \bibinfo{year}{2019}\natexlab{}.
\newblock \showarticletitle{Towards Automatic Face-to-Face Translation}. In
  \bibinfo{booktitle}{\emph{Proceedings of the 27th ACM International
  Conference on Multimedia}}. ACM, \bibinfo{pages}{1428--1436}.
\newblock


\bibitem[\protect\citeauthoryear{Kumar, Sotelo, Kumar, de~Br{\'e}bisson, and
  Bengio}{Kumar et~al\mbox{.}}{2017}]%
        {kumar2017obamanet}
\bibfield{author}{\bibinfo{person}{Rithesh Kumar}, \bibinfo{person}{Jose
  Sotelo}, \bibinfo{person}{Kundan Kumar}, \bibinfo{person}{Alexandre de
  Br{\'e}bisson}, {and} \bibinfo{person}{Yoshua Bengio}.}
  \bibinfo{year}{2017}\natexlab{}.
\newblock \showarticletitle{Obamanet: Photo-realistic lip-sync from text}.
\newblock \bibinfo{journal}{\emph{arXiv preprint arXiv:1801.01442}}
  (\bibinfo{year}{2017}).
\newblock


\bibitem[\protect\citeauthoryear{Maas, Hannun, and Ng}{Maas
  et~al\mbox{.}}{2013}]%
        {maas2013rectifier}
\bibfield{author}{\bibinfo{person}{Andrew~L Maas}, \bibinfo{person}{Awni~Y
  Hannun}, {and} \bibinfo{person}{Andrew~Y Ng}.}
  \bibinfo{year}{2013}\natexlab{}.
\newblock \showarticletitle{Rectifier nonlinearities improve neural network
  acoustic models}. In \bibinfo{booktitle}{\emph{Proc. icml}},
  Vol.~\bibinfo{volume}{30}. \bibinfo{pages}{3}.
\newblock


\bibitem[\protect\citeauthoryear{NPD}{NPD}{2016}]%
        {videocalling}
\bibfield{author}{\bibinfo{person}{NPD}.} \bibinfo{year}{2016}\natexlab{}.
\newblock \bibinfo{booktitle}{\emph{52 Percent of Millennial Smartphone Owners
  Use their Device for Video Calling, According to The NPD Group}}.
\newblock
\urldef\tempurl%
\url{https://www.npd.com/wps/portal/npd/us/news/press-releases/2016/52-percent-of-millennial-smartphone-owners-use-their-device-for-video-calling-according-to-the-npd-group/}
\showURL{%
\tempurl}


\bibitem[\protect\citeauthoryear{Suwajanakorn, Seitz, and
  Kemelmacher-Shlizerman}{Suwajanakorn et~al\mbox{.}}{2017}]%
        {suwajanakorn2017synthesizing}
\bibfield{author}{\bibinfo{person}{Supasorn Suwajanakorn},
  \bibinfo{person}{Steven~M Seitz}, {and} \bibinfo{person}{Ira
  Kemelmacher-Shlizerman}.} \bibinfo{year}{2017}\natexlab{}.
\newblock \showarticletitle{Synthesizing obama: learning lip sync from audio}.
\newblock \bibinfo{journal}{\emph{ACM Transactions on Graphics (TOG)}}
  \bibinfo{volume}{36}, \bibinfo{number}{4} (\bibinfo{year}{2017}),
  \bibinfo{pages}{95}.
\newblock


\bibitem[\protect\citeauthoryear{Thies, Elgharib, Tewari, Theobalt, and
  Nie{\ss}ner}{Thies et~al\mbox{.}}{2019}]%
        {thies2019neural}
\bibfield{author}{\bibinfo{person}{Justus Thies}, \bibinfo{person}{Mohamed
  Elgharib}, \bibinfo{person}{Ayush Tewari}, \bibinfo{person}{Christian
  Theobalt}, {and} \bibinfo{person}{Matthias Nie{\ss}ner}.}
  \bibinfo{year}{2019}\natexlab{}.
\newblock \showarticletitle{Neural Voice Puppetry: Audio-driven Facial
  Reenactment}.
\newblock \bibinfo{journal}{\emph{arXiv preprint arXiv:1912.05566}}
  (\bibinfo{year}{2019}).
\newblock


\bibitem[\protect\citeauthoryear{Tolosana, Vera-Rodriguez, Fierrez, Morales,
  and Ortega-Garcia}{Tolosana et~al\mbox{.}}{2020}]%
        {tolosana2020deepfakes}
\bibfield{author}{\bibinfo{person}{Ruben Tolosana}, \bibinfo{person}{Ruben
  Vera-Rodriguez}, \bibinfo{person}{Julian Fierrez}, \bibinfo{person}{Aythami
  Morales}, {and} \bibinfo{person}{Javier Ortega-Garcia}.}
  \bibinfo{year}{2020}\natexlab{}.
\newblock \bibinfo{title}{DeepFakes and Beyond: A Survey of Face Manipulation
  and Fake Detection}.
\newblock
\newblock
\showeprint[arxiv]{2001.00179}~[cs.CV]


\bibitem[\protect\citeauthoryear{Tursman, George, Kamara, and Tompkin}{Tursman
  et~al\mbox{.}}{2020}]%
        {Tursman_2020_CVPR_Workshops}
\bibfield{author}{\bibinfo{person}{Eleanor Tursman}, \bibinfo{person}{Marilyn
  George}, \bibinfo{person}{Seny Kamara}, {and} \bibinfo{person}{James
  Tompkin}.} \bibinfo{year}{2020}\natexlab{}.
\newblock \showarticletitle{Towards Untrusted Social Video Verification to
  Combat Deepfakes via Face Geometry Consistency}. In
  \bibinfo{booktitle}{\emph{Proceedings of the IEEE/CVF Conference on Computer
  Vision and Pattern Recognition (CVPR) Workshops}}.
\newblock


\bibitem[\protect\citeauthoryear{Vougioukas, Petridis, and Pantic}{Vougioukas
  et~al\mbox{.}}{2019}]%
        {vougioukas2019realistic}
\bibfield{author}{\bibinfo{person}{Konstantinos Vougioukas},
  \bibinfo{person}{Stavros Petridis}, {and} \bibinfo{person}{Maja Pantic}.}
  \bibinfo{year}{2019}\natexlab{}.
\newblock \showarticletitle{Realistic speech-driven facial animation with
  gans}.
\newblock \bibinfo{journal}{\emph{International Journal of Computer Vision}}
  (\bibinfo{year}{2019}), \bibinfo{pages}{1--16}.
\newblock


\bibitem[\protect\citeauthoryear{Wang, Bovik, Sheikh, Simoncelli,
  et~al\mbox{.}}{Wang et~al\mbox{.}}{2004}]%
        {wang2004image}
\bibfield{author}{\bibinfo{person}{Zhou Wang}, \bibinfo{person}{Alan~C Bovik},
  \bibinfo{person}{Hamid~R Sheikh}, \bibinfo{person}{Eero~P Simoncelli},
  {et~al\mbox{.}}} \bibinfo{year}{2004}\natexlab{}.
\newblock \showarticletitle{Image quality assessment: from error visibility to
  structural similarity}.
\newblock \bibinfo{journal}{\emph{IEEE transactions on image processing}}
  \bibinfo{volume}{13}, \bibinfo{number}{4} (\bibinfo{year}{2004}),
  \bibinfo{pages}{600--612}.
\newblock


\bibitem[\protect\citeauthoryear{Zhou, Liu, Liu, Luo, and Wang}{Zhou
  et~al\mbox{.}}{2018}]%
        {zhou2018talking}
\bibfield{author}{\bibinfo{person}{Hang Zhou}, \bibinfo{person}{Yu Liu},
  \bibinfo{person}{Ziwei Liu}, \bibinfo{person}{Ping Luo}, {and}
  \bibinfo{person}{Xiaogang Wang}.} \bibinfo{year}{2018}\natexlab{}.
\newblock \showarticletitle{Talking Face Generation by Adversarially
  Disentangled Audio-Visual Representation}.
\newblock \bibinfo{journal}{\emph{arXiv preprint arXiv:1807.07860}}
  (\bibinfo{year}{2018}).
\newblock


\end{thebibliography}

\end{document}